\documentclass[letterpaper, 10 pt, conference]{ieeeconf}  

\IEEEoverridecommandlockouts                              
\usepackage[table]{xcolor}
\usepackage{graphicx}				
\usepackage[labelformat=simple]{subcaption}	
\usepackage{amssymb}
\usepackage{amsmath}
\usepackage{mathtools}
\usepackage{listings}
\usepackage{url}
\usepackage{epstopdf}
\usepackage{array}
\usepackage{tikz}
\usepackage{amsmath}
\usepackage{algorithm}
\usepackage{algpseudocode}
\usepackage{multirow}
\usepackage{flexisym}
\usepackage{amsmath}
\usepackage{array}
\usepackage{bbold}
\usepackage{subfiles}
\usepackage{cite}
\usepackage{gensymb}
\usepackage{soul}
\usepackage[normalem]{ulem}
\usepackage[margin=0pt,font=small,labelsep=period]{caption}
\newcommand{\vtxt}[1]{\textcolor{orange}{{}}}
\definecolor{cerise}{rgb}{0.85, 0.2, 0.39}

\definecolor{green}{RGB}{0,255,0}

\newcommand{\SGNet}{\texttt{PtPNet\,}}
\newcommand{\robot}{\mathcal{R}}     
\newcommand{\camera}{\mathcal{C}}     
\newcommand{\camerap}{\mathcal{{C^{\prime}}}}     
\newcommand{\shoe}{\mathcal{S}}   
\newcommand{\hook}{\mathcal{E}}   
\newcommand{\rot}{\mathbf{R}}   
\newcommand{\reals}{\mathbb{R}} 
\newcommand{\rbt}[2]{{{^#1} T_{#2}}} 
\newcommand{\waypoint}{w}       
\newcommand{\trajectory}{W}     
\newcommand{\nwaypoints}{N}     

\title{Pixels to Plans: Learning Non-Prehensile Manipulation \\ by Imitating a Planner}

\author{Tarik Tosun$^*$, Eric Mitchell$^*$, Ben Eisner, Jinwook Huh, Bhoram Lee, Daewon Lee,\\ Volkan Isler, H. Sebastian Seung, and Daniel Lee
\thanks{*These authors contributed equally to this work}
\thanks{All authors are with the Samsung AI Center NY, 123 West 18th Street, New York, New York 10011}%
}

\begin{document}

\maketitle
\thispagestyle{empty}
\pagestyle{empty}

\begin{abstract}
We present a novel method enabling robots to quickly learn to manipulate objects by leveraging a motion planner to generate ``expert'' training trajectories from a small amount of human-labeled data. In contrast to the traditional sense-plan-act cycle, we propose a deep learning architecture and training regimen called \SGNet that can estimate effective end-effector trajectories for manipulation directly from a single RGB-D image of an object. Additionally, we present a data collection and augmentation pipeline that enables the automatic generation of large numbers (millions) of training image and trajectory examples with almost no human labeling effort.

We demonstrate our approach in a non-prehensile tool-based manipulation task, specifically picking up shoes with a hook. In hardware experiments, \SGNet generates motion plans (open-loop trajectories) that reliably (89\% success  over  189  trials)  pick  up  four  very  different shoes from a range of positions and orientations, and reliably picks up a shoe it has never seen before.  Compared with a traditional sense-plan-act paradigm, our system has the advantages of operating on sparse information (single  RGB-D  frame), producing  high-quality trajectories much faster than the “expert” planner (300ms versus several seconds), and generalizing effectively to previously unseen shoes.

%

\end{abstract}

\section{Introduction}
\begin{figure}[t]
    \centering
    \includegraphics[width=0.9\columnwidth]{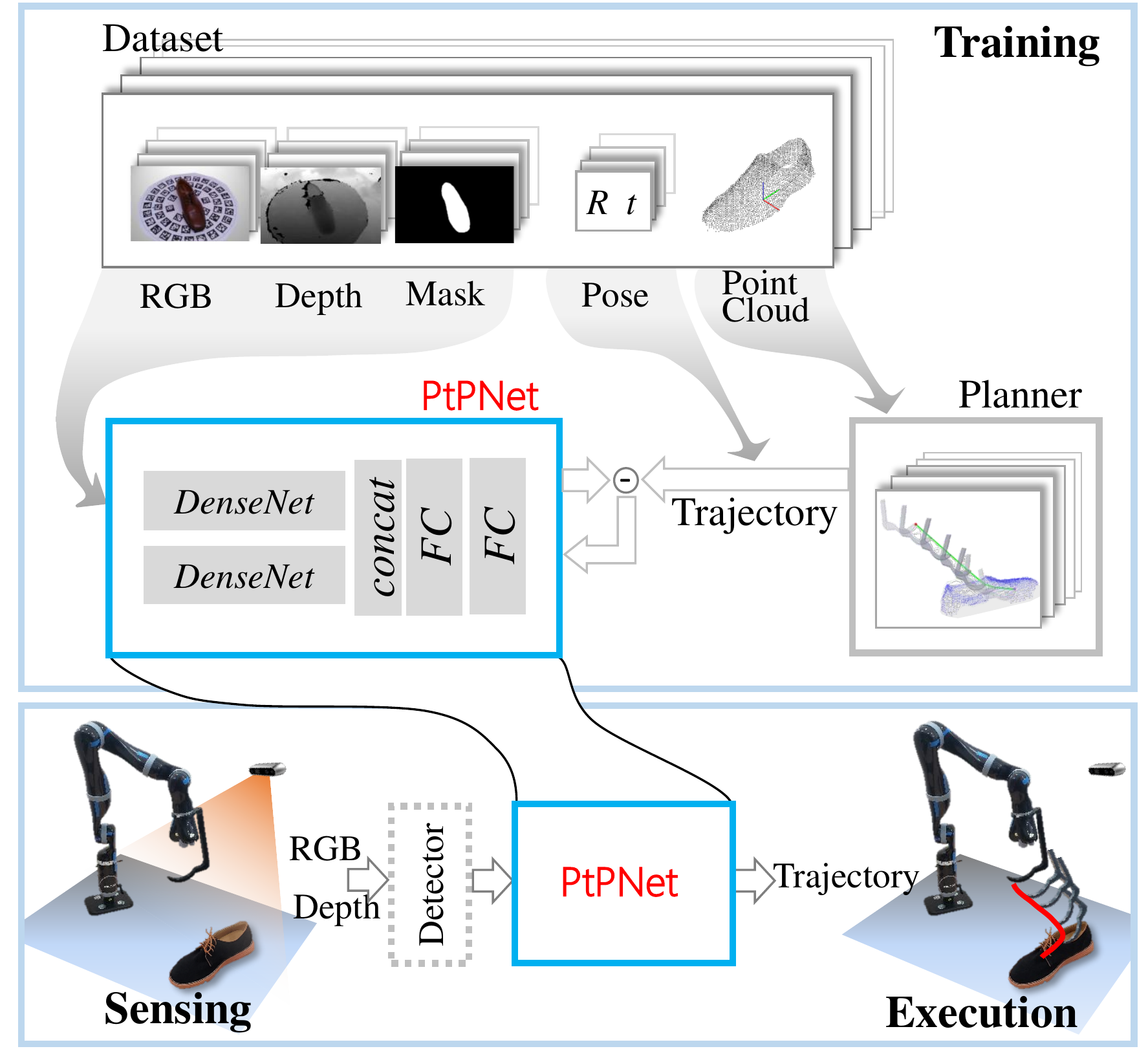}
    \caption{Conceptual System Overview. During training, we leverage a high-quality point cloud model of a shoe and an RRT motion planner to generate ground truth trajectories for each training example used to train \SGNet. At inference time, we rely on only a single RGB-D camera sample and an attention mechanism/object detector to directly output an end-effector trajectory in the camera frame.}
    \label{fig:overview_flowchart}
\end{figure}

Despite many advances in robotic manipulation in industrial settings, manipulating general objects in unstructured environments remains challenging.
The traditional approach for manipulation relies on the \emph{sense-plan-act} paradigm which decouples these three components \cite{chitta2012perception}. A common example comprises of a camera module that captures camera input and processes it generate an intermediate geometric representation of the object to be manipulated, a trajectory planner which generates a path based on this representation, and a path-following controller that executes the path.

Decoupling these components allows for independent progress in complementary areas. For example, recent advances in object detection and segmentation in images can be directly incorporated in the sensing module.
Similarly, state-of-the art planners can be used for generating trajectories in high-dimensional configuration spaces. However, hand-designing the interface between these components can introduce brittleness at the system level. For example, even though the planner can effectively generate a trajectory when given a complete three-dimensional model and the pose of the object to be manipulated, it might be too difficult for the sensing system to generate a precise 3D model and pose for a given input image. 

Recent ``pixels-to-actions'' methods have shown promise in addressing these challenges posed by rigid interfaces. Rather than requiring an explicit intermediate representation, ``pixels-to-actions'' methods estimate effective actions (in the form of joint angles or torques) directly from raw sensor data, without any explicit intermediate state \cite{DBLP:journals/corr/LevineFDA15, DBLP:journals/corr/LevinePKQ16}. 
However, current pixels-to-actions techniques often suffer from both high sample complexity and brittleness in the presence of deviations from the learning environment, which are particularly significant in robotic applications.  \vtxt{please check this:} So far, these methods have mostly been used in very task-specific environments because learning the dynamics of the task at the controller level and discovering appropriate actions requires numerous training examples. The direct coupling of sensor input to controller actions may be too restrictive and leads to bad generalization performance.

In this paper, we present a new approach in which we train a deep neural network called \SGNet to generate a motion plan (represented as a sequence of trajectory waypoints) from a single RGB-D image.
We demonstrate this approach in the context of a tool-based manipulation task, specifically picking up shoes with a hook.  Manipulation with a passive tool presents a challenging motion planning problem, because it requires moving the tool through a potentially complex sequence of positions and orientations with respect to the object being manipulated, as opposed to selecting a single grasp pose for a gripper.  The range of shapes of the shoes in our training and test sets require a variety of qualitatively different hooking trajectories to manipulate them all effectively.

Provided with only partial information about the pose and geometry of a shoe (in the form of a single RGB-D image), \SGNet is trained to closely replicate example trajectories generated by an ``expert'' motion planner that has access to detailed information about the pose and geometry of the shoe.
Core to our training paradigm is a dataset of 3D-scanned shoes that registers many individual RGB-D views to a single dense point cloud for each shoe, a trajectory generation framework that employs a motion planner to generate example trajectories from shoe point clouds, and a robust data augmentation procedure that automatically generates millions of data samples in the form of input image/ground-truth trajectory pairs over the course of training.

\SGNet's training corpus is based on a relatively small number of 7335 images of 34 shoes, generated by an automated 3d capture system. Using the augmentation of procedure described in Section~\ref{sec:augmentation} we can generate thousands of new images (and matching trajectories) for each original image and thereby effectively increase the training size to millions of images over the course of training.
In hardware experiments, we demonstrate that the network successfully generates open-loop trajectories that reliably (89\% success over 189 trials) pick up four very different shoes from a range of positions and orientations within the camera view, and generalizes to reliably pick up a shoe it has never seen before.  Our results demonstrate that \SGNet has learned to infer from a single RGB-D image \textit{what kind} of shoe it is seeing, \textit{where} the shoe is with respect to the camera, and ultimately \textit{how to move the hook} to capture the shoe.  Compared with a traditional sense-plan-act paradigm, it has the advantages of operating on \textbf{sparse information} in the partially-observed setting (single RGB-D frame rather than a complete 3D model), producing high-quality trajectories much \textbf{faster} than the ``expert'' planner (300ms versus several seconds), and effective \textbf{generalization} to shoes it has never seen before (for which dense 3D information is not available).

Compared to many `pixels-to-actions' paradigms, our method achieves robust manipulation without the need for any training on a real robot and with only very few human-labeled annotations. Our proposed system can also be implemented (as we do) in a way that generalizes across different robotic and camera hardware and conditions, making it desirable for use as a general-purpose manipulation learning method.

%
\begin{figure}
    \centering
    \includegraphics[height=2in]{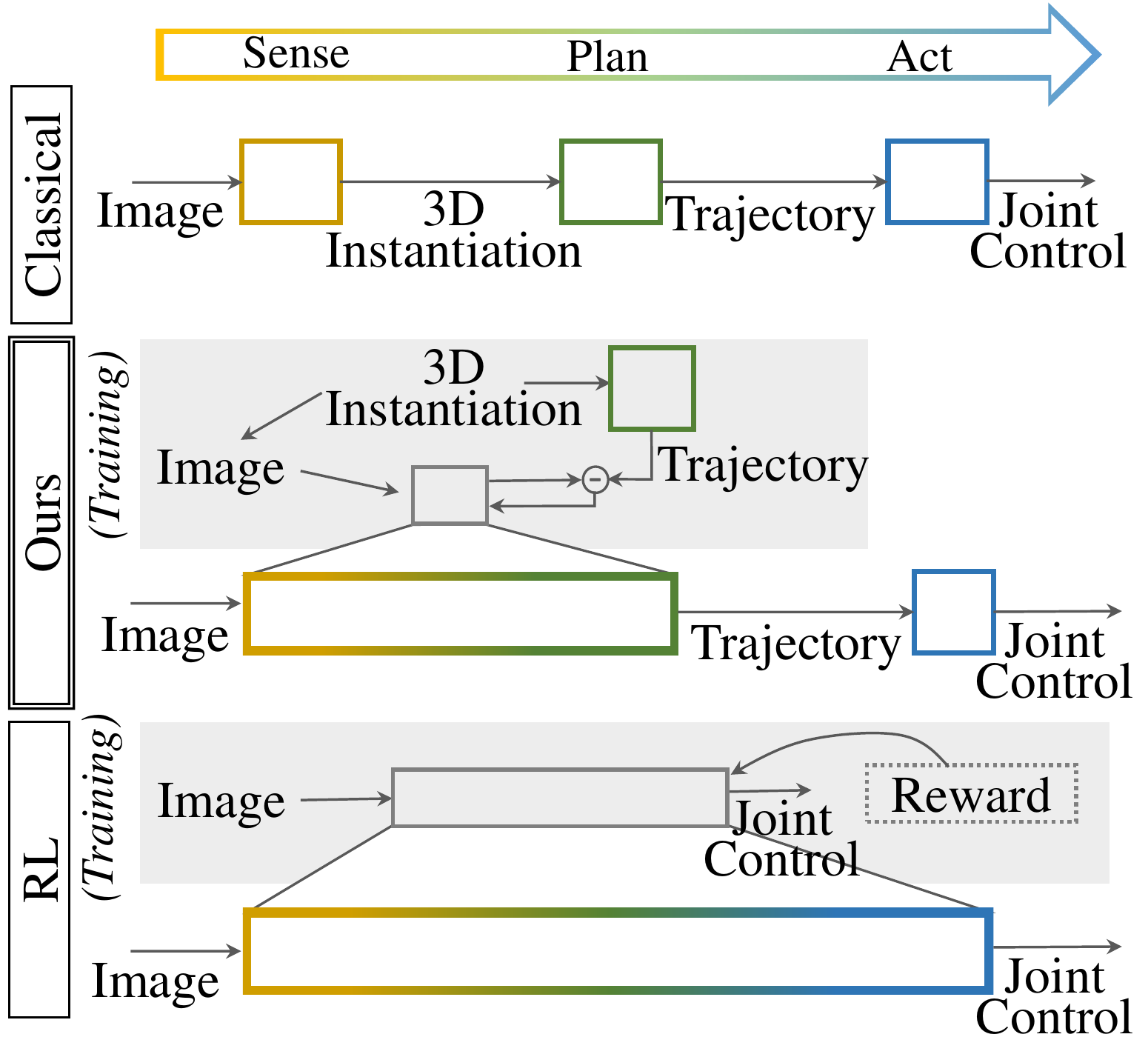}
    \caption{Conceptual comparison of our method (middle) with the classical sense-plan-act paradigm (top) and end-to-end reinforcement learning methods (bottom). Our method mitigates the need for a structured representation of the scene at inference time (as in the classical paradigm), but produces a generic plan that any controller can follow instead of directly outputting actions for a specific controller.
   }
    \label{fig:rl-classical-comparison}
\end{figure}

\section{Related Work}

Several approaches have been proposed for applying deep learning to the problem of robotic manipulation. Levine et al. \cite{DBLP:journals/corr/LevineFDA15} trained a convolutional net to map raw images and joint angles to joint torques. The net was trained through policy search guided by a linear-Gaussian controller. The work demonstrated that a pixels-to-torques mapping could be learned, but was limited to manipulation tasks that could be solved by a linear-Gaussian controller.

In later work, Levine et al.\cite{DBLP:journals/corr/LevinePKQ16} trained a convolutional net to predict grasp success given a raw image and a one-step motion plan. At run time, the motion plan at each time step was chosen to maximize the net's prediction of grasp success. This approach was formalized as a kind of Q-learning \cite{kalashnikov2018qt}, and required a large number of manipulation trials to learn an accurate predictor of grasp success. The trials were generated by building an``arm farm'' with many robotic arms learning to manipulate in parallel.

Zhang et al.\cite{zhang2018deep} trained a convolutional net to generate a motion plan from raw images and end-effector pose. In this deep imitation learning approach, the target values for the motion plan were provided by virtual reality teleoperation of the manipulator by a human teacher. A potential disadvantage is that collection of large amounts of training data may require significant labor by human teachers. 

Here we propose an alternative deep imitation learning approach in which the targets for supervised training are provided by a trajectory planner rather than a human teacher. A similar approach has been applied to autonomous driving \cite{pan2018agile}, but learning to imitate a planner is novel in the domain of manipulation as far as we know. Contemporary planning algorithms such as RRT\cite{lavalle2001randomized} are quite powerful; however, these algorithms may be slow. As our empirical results will show, training a convolutional net to imitate the planner yields performance that is fast, accurate, and generalizes to objects that were not seen during training. Furthermore the net operates with sparser information (single view RGB-D) than the planner (full 3D collision geometry).

The prior works mentioned above were used to train closed-loop systems that use sensory feedback while generating motions. In the present work, we train an open-loop system to generate a motion plan using sensory information from a single image. This was done for simplicity; extensions to closed-loop manipulation control will be sketched in the Discussion.

\section{Problem Statement and Overview}
\subsection{Problem Statement}
We consider a setup which consists of a manipulator (whose kinematics are known) and an external fixed RGB-D camera. 
Throughout the paper, $\robot$ denotes the base frame of the manipulator, $\hook$ denotes the frame of the robot end-effector (which in this case is a hook tool), $\camera$ is the camera frame and $\shoe$ is the shoe frame. The pose of the camera with respect to the robot $\rbt{\robot}{\camera}$ is assumed to be known. Given a single RGB-D image of the shoe from the camera, the objective is for the robot to pick up the shoe using the hook.  Picking is considered successful if at the end of the robot's motion, the shoe has been lifted completely off the table and hangs stably on the hook, and the shoe and hook are not damaged during the motion.

\subsection{Method Overview}
Our method trains a neural network that
accepts as input a single RGB-D image of a shoe and outputs a \textit{camera-frame end-effector trajectory} $^\camera \trajectory = \{ \waypoint_1, \waypoint_2 \ldots \waypoint_\nwaypoints \}$ consisting of $\nwaypoints$ waypoints, where each waypoint $\waypoint_k = \rbt{\camera}{\hook_k}$ defines a pose of the end-effector with respect to the camera.  Since the camera-to-robot calibration $\rbt{\robot}{\camera}$ is assumed to be known, this camera-frame trajectory can be transformed into the robot frame for execution by the robot: $^\robot \trajectory = \rbt{\robot}{\camera}~{^\camera}\trajectory$.   
  
Our network is trained via imitation learning to closely replicate example trajectories generated by an ``expert'' motion planner that has access to detailed information about the pose and geometry of the shoe.  Figure~\ref{fig:overview_flowchart} shows an overview of our training framework.  Our dataset consists of dense 3D point cloud models, RGB-D images, masks, and poses of real shoes, generated using a data capture system consisting of a turntable and three Intel Realsense RGB-D cameras (Fig.~\ref{fig:data_capture}).  
For each shoe, an ``expert'' end-effector trajectory $^\camera \trajectory^*$ is generated by an RRT motion planner that has access to a 3D model of the end-effector tool, a dense point cloud of the shoe, a desired goal position for the hook within the shoe, and the pose of the shoe in the camera frame.

The images from the capture system coupled with these trajectories provide a core set of training data.
The network is provided with an RGB-D image of the shoe as input, and produces a camera-frame end-effector trajectory $^\camera \trajectory$ as output, which is then compared with the expert trajectory $^\camera \trajectory^*$, and the network is iteratively trained to minimize a loss function which measures the similarity of the two trajectories.

In Section~\ref{sec:methods}, we describe our neural network architecture, example trajectory generation process, and data collection and augmentation procedures in detail.  In Section~\ref{sec:experiments}, we present hardware experiments that benchmark our system in a shoe-picking task, characterize its ability to generalize across shoe poses, and test its ability manipulate shoes that were not in the training set.  In Seciton~\ref{sec:discussion}, we discuss lessons learned, and conclude.
\section{Methods}\label{sec:methods}
\subsection{Neural Network Architecture and Training}
%
\subsubsection{Architecture}

\newcommand{\Rotf}{\text{Rot}}
\newcommand{\Transf}{\text{Trans}}
\newcommand{\cw}{\waypoint^*}
\newcommand{\ct}{^\camera\trajectory}
\newcommand{\cwh}{\hat{\waypoint}}
\newcommand{\cth}{\hat{^\camera\trajectory}}
\label{section:arch}


%
\begin{figure}
    \centering
    \includegraphics[width=1.0\columnwidth]{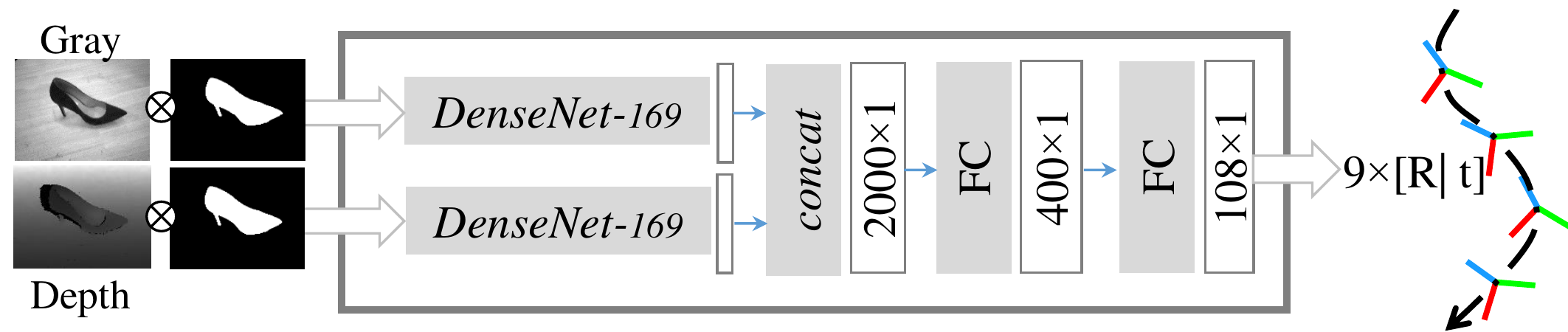}
    \caption{\SGNet Architecture. From an input consisting of an $(I^g,I^d)$ pair, corresponding to the grayscale and depth images from an RGB-D camera, we compute $^\camera\hat{\trajectory}_\hook = \{\hat{\waypoint}_1,\hat{\waypoint}_2,...,\hat{\waypoint}_N\}$. $I^g$ and $I^d$ are fed into two separate, identical DenseNet-169 instantiations, after which they are combined into several fully-connected layers and then output as a trajectory.
    }
    \label{fig:network_architecture}
\end{figure}
The fundamental building block of our neural network architecture is the successful DenseNet paradigm introduced in \cite{huang2017densely}, specifically the variant DenseNet-169\footnote{Source code of the DenseNet-169 building block is located here: \texttt{https://pytorch.org/docs/1.0.0/\_modules/torchvision/ \\ models/densenet.html\#densenet169}} (a particular instantiation of the DenseNet concept) proposed in \cite{huang2017densely} and implemented in the deep learning framework PyTorch \cite{paszke2017automatic}. DenseNet-169 contains 168 convolutional layers and a single fully connected layer following the learned features. The general DenseNet architecture was chosen as the backbone of our network because it has been shown to be a powerful, parameter-efficient architecture that learns quickly, yielding state-of-the-art results on a variety of computer vision tasks including classification, segmentation, and real-time object detection on mobile devices \cite{DBLP:journals/corr/HuangCLWMW17} \cite{DBLP:journals/corr/JegouDVRB16} \cite{DBLP:journals/corr/abs-1804-06882}.

The network architecture used in this paper, which we call \SGNet, uses two separate DenseNet-169 modules to process the grayscale and depth measurements from the camera in separate streams, which are ultimately joined by a sequence of two fully connected layers at the end of the network. Weights are \textit{not} shared across these streams as in some `Siamese'' architectures \cite{Koch2015SiameseNN}. A diagram of this architecture is given in Figure \ref{fig:network_architecture}. 

The input to the network is a pair of aligned grayscale and depth images $(I^g, I^d)$. The network assumes that both images have been foreground-masked, that is, non-shoe pixels and depth values have been set to zero. Several possible techniques for acquiring such masks exist, including simply estimating a plane and filtering depth points or even training a separate convolutional neural network for object segmentation. We implemented and tested both strategies for this work, and both were effective. In our experiments, we performed shoe segmentation with a convolutional neural network based on UNet \cite{DBLP:journals/corr/RonnebergerFB15} that was trained on a small subset (roughly 15\%) of our trajectory dataset; we chose to use a multi-scale convolutional network over other segmentation techniques (e.g. plane subtraction from depth image) because its output was substantially more robust to noise from the RGB-D camera, and is robust to non-shoe objects in the frame.

In the forward pass, each image is first processed separately by one of two DenseNet-169 networks to extract 1000 features each, for a total of 2000 features. Two more hidden layers, coupled with the ReLU activation function \cite{Nair:2010:RLU:3104322.3104425}, compute `mixed' features, and a final output layer directly regresses the trajectory plan estimate $\cth=\{\cwh_i:i\in 1,...,n\}$ as a single vector $\hat{t} \in \reals^{(12\nwaypoints)}$ for $\nwaypoints$ trajectory waypoints. In this paper, we use $\nwaypoints=9$, but our framework is general to other trajectory lengths. This output vector is interpreted as a sequence of sub-vectors $\hat{t}_i \in \reals^{12}$, each corresponding to a homogeneous transform $\hat{w}_i=\hat{\rbt{\camera}{\hook}}$ from the camera coordinate frame to the desired coordinate frame of the end effector of the robot at waypoint $i$. For each waypoint vector $t_i$, the first 3 values $(t_i^1, t_i^2, t_i^3)$ represent the desired $(x,y,z)$ position of the robot's end effector in the camera's coordinate frame. The last 9 values represent a serialized 3D rotation matrix giving the relative orientation of the robot's end effector relative to the camera for that waypoint. Learning end-effector trajectories in the camera frame decouples the learned solution to the task from the position of the robot; as long as the camera-to-robot calibration $\rbt{\robot}{\camera}$ is known, we can use the trained network with arbitrary robot position and camera positions with \textbf{no} retraining. Provided they have similar dexterous workspaces and identical end-effectors, the method should also generalize without retraining to an entirely different robots (there is no notion of a robot during training, only an end-effector trajectory).
\subsubsection{Training}

During training, the network is presented with input/output pairs $((I^g, I^d), {\ct^*})$, where $I^g$ and $I^d$ are grayscale and depth images from the same scene, respectively, and $\ct^*$ is the corresponding ground-truth trajectory generated by the RRT motion planner. Ground-truth trajectories are serialized as a single vector, as described in Section \ref{section:arch}.

For each input $(I^g, I^d)$ seen during training, the network makes a prediction $\cth=\{\cwh_i:i\in 1,...,N\}$ and a loss is computed from the ground truth trajectory $\ct$. The loss for the trajectory estimate $\cth$ is the weighted sum of the individual trajectory waypoint losses:
\begin{equation}
    L_{traj}(\cwh, \cw) = \sum_{i=1}^{n}\alpha_i\ell(\cwh_i, \cw_i)
\end{equation}

This trajectory waypoint loss, given its corresponding ground truth trajectory waypoint $\cw_i$, is computed by first decomposing each waypoint into its representative translation and rotation sub-components. For any 12 dimensional waypoint vector $\waypoint$, we define the functions $\Transf(\waypoint):\reals^{12} \rightarrow \reals^3$ and $\Rotf(\waypoint):\reals^{12} \rightarrow \reals^{3\times 3}$, which extract the position and rotation matrix of a given waypoint relative to the camera coordinate frame, respectively. Using this notation, the trajectory waypoint loss is given as:

\begin{equation}
    \ell(\cwh_i, \cw_i) = \lambda \ell_{T}(\cwh_i, \cw_i) + \gamma \ell_{R}(\cwh_i, \cw_i)
\end{equation}

where $\ell_{T}(\cwh_i, \cw_i)$ is the squared Euclidean distance loss:

\begin{equation}
    \ell_{T}(\cwh_i, \cw_i) = ||\Transf(\cwh_i) - \Transf(\cw_i)||^2
\end{equation}

and $\ell_{R}(\cwh_i, \cw_i)$ is the squared deviation of the product of the predicted rotation matrix and the transpose of the ground truth rotation matrix:

\begin{equation}
    \ell_{R}(\cwh_i, \cw_i) = ||\Rotf(\cwh_i)\Rotf(\cw_i)^T - I||^2
\end{equation}

In this work, we use $\lambda = \gamma = 1$ and $\alpha_i = 1$ for all $i$; that is, we weight the rotation matrix deviation loss and waypoint coordinate loss equally and weight all individual waypoint losses equally within a trajectory. We train with the Adam optimizer \cite{DBLP:journals/corr/KingmaB14} with learning rate 1e-4, batch size 64, and weight decay coefficient 1e-4 for 1000 epochs on 4 NVIDIA 1080 Ti GPUs.

\subsection{Dataset}
\begin{figure}[t]
    \centering
    \includegraphics[width=0.8\columnwidth]{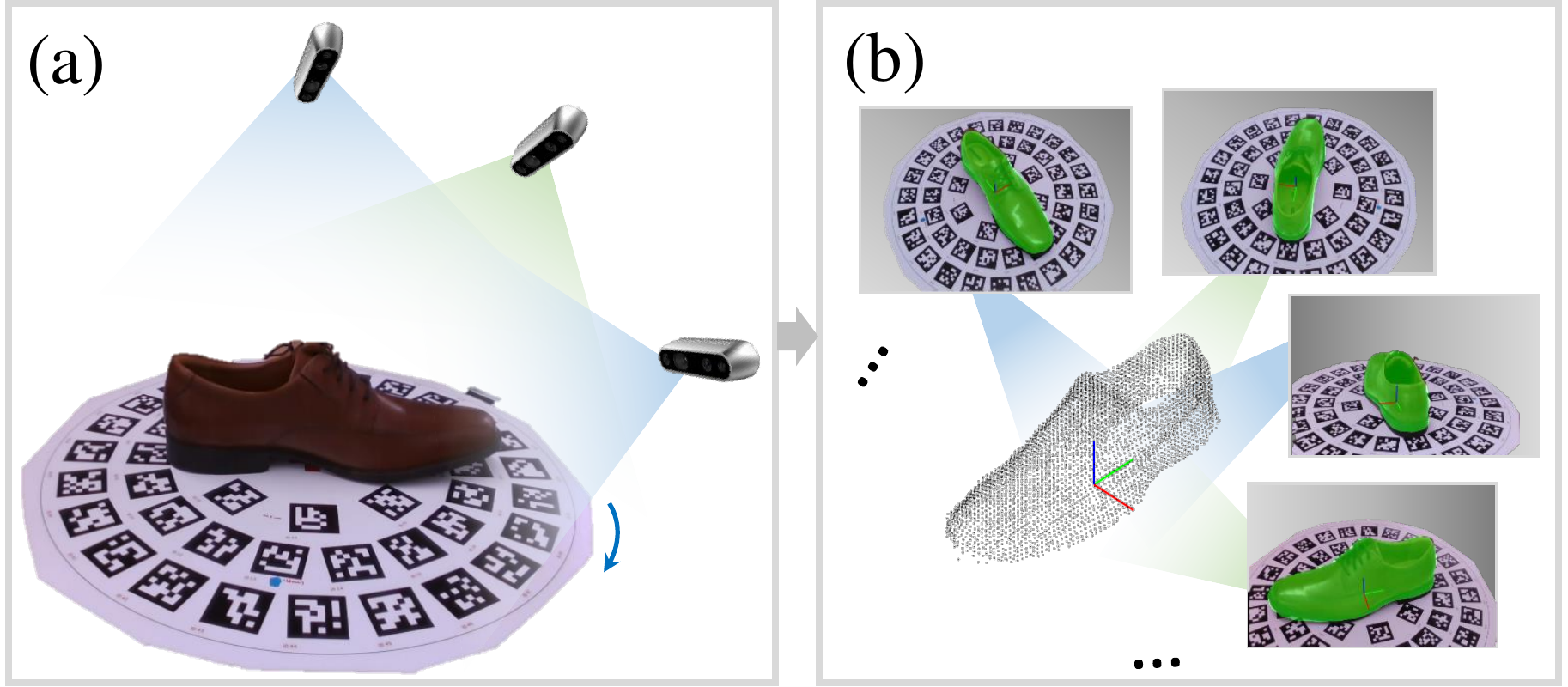}
    \caption{Training dataset capture procedure. (a) shows the physical collection configuration, with a single shoe mounted on a turntable marked with AprilTags, and three cameras placed at different poses around the turntable. The turntable is rotated a full 360\degree, with RGB-D images captured at 5\degree intervals. (b) shows the computational procedure used to extract a complete point cloud of the shoe by synthesizing multiple RGB-D views into a single model. }
    \label{fig:data_capture}
\end{figure}

%
In order to train \SGNet, we have compiled a dataset comprised of point cloud models, RGB-D images, masks, and poses of real shoes. The capture setup includes three low-cost RGB-D cameras (Intel RealSense D435\footnote{\url{https://realsense.intel.com/depth-camera/}}), a controllable turn-table, and an AprilTag\cite{olson2011apriltag} pattern board (Fig.\ref{fig:data_capture}(a)). 

The cameras capture images of the exact same scene from different camera poses relative to the target object when the turn-table stops at a certain interval while rotating. After the images are collected, the camera pose of each frame is computed using AprilTags. The depth point cloud of the target is obtained by removing all other points except the target area above the pattern board and by filtering using a voxel-based simplification method. The point cloud can be re-projected onto each RGB image to generate the mask and its pose can be computed with respect to the corresponding camera frame  (Fig.\ref{fig:data_capture}(b)). In this way, each model includes 216 views (5 deg interval $\times$ 3 cameras) recorded in RGB, IR, and depth, as well as the shoe pose for each view. At the moment, the dataset include 45 shoe instance models, 34 of which were used for training \SGNet. 

Note that we do not impose a clean surface structure to build a mesh model which is difficult to obtain dynamically. This approach enables gathering images of real objects with masks and poses without the need for 3D surface models. The capture and labeling process is automated without any manual annotation.

%

%
\subsection{Example Trajectories}
\subsubsection{Trajectory Representation}
We define a \textit{shoe-frame trajectory} $^\shoe \trajectory$ as a set of $\nwaypoints$ waypoints of the end effector with respect to the shoe frame. Let a waypoint $\waypoint_k$ be $[x_k,y_k,z_k, \phi_k, \theta_k, \psi_k]$ in $SE(3)$ space and $\waypoint_k \in {^\shoe \trajectory} \: \textrm{for} \: k \in {1,\dots, N}$, where $[x_k,y_k,z_k]$ is a position, and $[\phi_k, \theta_k, \psi_k]$ are Euler angles of the hook. For trajectory generation, we use the Euler angle representation of rotations because it is stable and relatively computationally efficient. For training of \SGNet, we represent $SO(3)$ as a rotation matrix $\rot_k$ since Euler angle representation has an ambiguity due to multiple parameter values for the same rotation representation \cite{kuffner2004effective}. In addition, we fix the number of waypoints $\nwaypoints$ as $9$ in this paper. The complete trajectory is thus represented by a sequence of (9) 12-dimensional keypoints, or a single 108-dimensional vector.

In order to train \SGNet, we need ground truth end-effector trajectories \textit{in the camera frame} for each grayscale/depth image pair in our dataset. However, because we collect the shoe to camera transform $\rbt{\camera}{\shoe}$ for each sample of our dataset, we only need to generate a single trajectory for each \textit{shoe (N=34)} rather than for each \textit{shoe image (N=7335)}. Using a motion planner, we generate a \textbf{single} example trajectory $^\shoe W$ for each shoe in the shoe's own coordinate frame. For each image pair and corresponding shoe-to-camera transform $((I^g,I^d),\rbt{\camera}{\shoe})$ in our dataset, we can then generate an appropriate camera-frame trajectory ${^\camera \trajectory}$ by simply transforming ${^\shoe \trajectory}$ into the camera frame: $^\camera W = \rbt{\camera}{\shoe}{^\shoe W}$. Thus, our dataset is generated from exactly 34 human-labeled annotations.

%

\subsubsection{RRT Trajectory Generation}

\begin{figure}[t]
    \centering
    \includegraphics[width=\columnwidth]{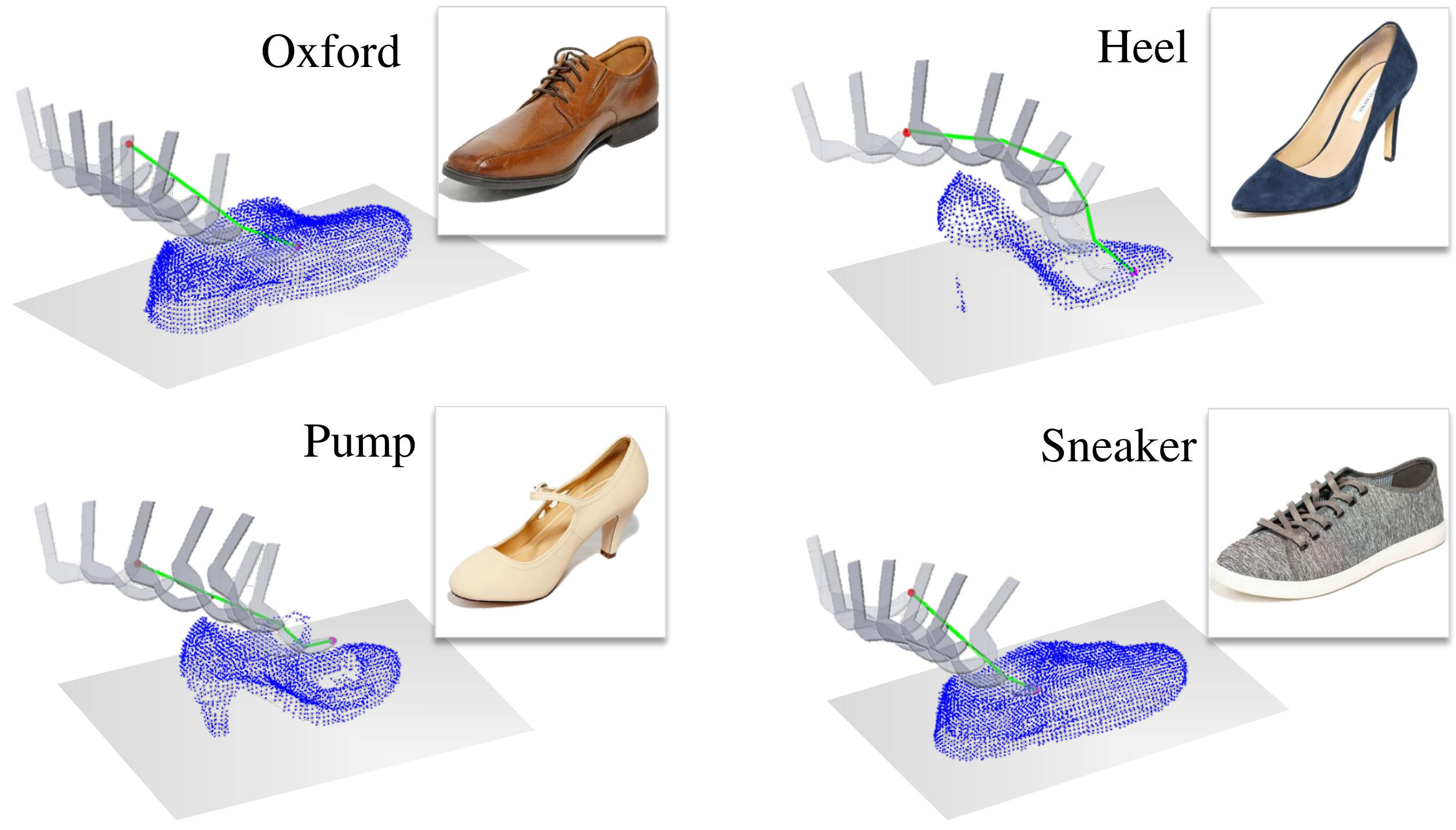}
    \caption{RRT planning from 3D point clouds. In this figure, we demonstrate qualitatively different hooking trajectories output by the RRT algorithm when run on shoes with different morphologies. Notice how the trajectories approach at different angles depending on what occlusions the shoes themselves introduce.}
    \label{fig_rrt_plan}
    \label{fig:rrt_plan}
\end{figure}

To achieve robust shoe hooking across a wide class of shoes (that vary significantly in color, texture, and shape), a robot must necessarily utilize distinct manipulation strategies for sufficiently distinct shoes. For example, the actions required to pick up a high heel are fundamentally different from those required to pick up a sneaker, as the shape and mass distributions are quite different. Fig. \ref{fig_rrt_plan} demonstrates the distinct trajectories required to successfully hook four different shoes. In this paper, we apply a sampling based planner to generate appropriate hooking trajectories for each shoe based on its point cloud. We define a goal pose inside a shoe manually for the sampling based planning, and we an RRT (Rapidly exploring Random Trees) motion planner to generates a trajectory from a fixed initial pose near the shoe to the specified target goal pose without colliding with the shoe point cloud. 
We employ the uniform sampling method of Euler angles and distance metric in \cite{kuffner2004effective} to effectively sample $SE(3)$. In addition, we use a bidirectional approach and 10\% goal biased sampling to improve performance.

Collision checks are computed by approximating the geometry of the hook end-effector as a point cloud, and comparing the  minimum distance between the end-effector point cloud and shoe point cloud. Once a path to the goal point is found, one additional waypoint is added directly above the goal point to lift the shoe. This waypoint is $20cm$ higher in the $z$ direction from the goal pose. 

 Sampling-based planners sometimes generate jerky and unnatural paths, so after planning the RRT we apply a path smoothing heuristic. Possible path smoothing techniques include the shortcut or spline algorithms \cite{hauser2010fast, geraerts2007creating}; we chose to apply the shortcut smoothing method due to its empirically determined effectiveness on our task and its low computational complexity. Functionally, this algorithm repeats 100 iterations in which it randomly chooses two configurations and linearly interpolates between the two if no collisions are detected. After smoothing the trajectory, we choose $9$ evenly-spaced waypoints from the trajectory to represent the ground truth trajectory for training.


\paragraph{sampling and distance metric}



%
\subsection{Data Augmentation}
\label{sec:augmentation}
\def\block(#1,#2)#3{\multicolumn{#2}{c}{\multirow{#1}{*}{$ #3 $}}}

When trained solely on raw images obtained by the acquisition system and their corresponding correct trajectories, \SGNet learns a trajectory-generating function that fails to produce successful trajectories when the test case deviates from the training conditions, e.g., when the shoe is not carefully centered in the image or the camera is closer or farther from the shoe than in training by more than a few centimeters. This is to be expected, since the raw dataset contains only images and point clouds of shoes in relatively homogeneous configurations, and thus the only trajectories the net is exposed to are trajectories that hook shoes at a specific distance from the camera and at the center of the image. To mitigate these geometric test-time limitations, we used the detailed 3-dimensional data collected for each shoe to generate new examples during training that span the entire field of view of the camera at a wide range of depths. 

First, we define a sample tuple as $(I^g, I^d, {^\camera \trajectory}, \rbt{\camera}{\shoe})$, corresponding to the grayscale image, depth image, ground truth trajectory of the tool in the camera frame, and pose of the shoe in the camera frame, $\camera$, respectively. Applying augmentation to a sample requires applying a single augmentation transformation $\rbt{\camerap}{\camera}$ to each attribute of the sample, which is interpreted as re-capturing the sample from a new camera pose $\camerap$. We achieve this by simulating a random rotation about the $x$ and $y$ axes of the camera frame, forming the rotation matrix $\rot_{\theta\phi} = \rot_y(\phi)\rot_x(\theta)$, as well as a random displacement in the direction of the camera frame $z$ axis, $\Delta z$. We bound these random parameters such that the all shoes remain within the camera's field of view. The intuition behind this camera rotation and $z$-displacement is that it will \textit{shift} and \textit{scale} the image of the shoe without deforming its appearance. 

We form the augmentation transformation $\rbt{\camerap}{\camera}$ as:

\[ 
    \rbt{\camerap}{\camera} = 
    \begin{bmatrix}
    \rot_{\theta\phi} & \boldsymbol{t}\\
      0 & 1\\
    \end{bmatrix},\quad \boldsymbol{t} = \begin{bmatrix} 0 & 0 & \Delta z \end{bmatrix} ^T 
\]

Augmenting $I^d$, $\rbt{\camera}{\hook}$, and $\rbt{\camera}{\shoe}$ thus involves applying a single homogeneous coordinate frame transformation (augmenting the depth image requires projecting the points with the inverse camera projection matrix, $K^{-1}$, applying the $\rbt{\camerap}{\camera}$, and then re-projecting with $K$). 

We approximate the application of this transformation to the image $I^g$ by approximating the scaling effect of the depth shift first and then applying the rotation as a shift in the image plane. To account for the depth offset $\Delta z$, we scale the image size by a factor of $\frac{z_0}{z_0 - \Delta z}$, where $z_0$ is the original $z$ position of the shoe frame in the camera frame $\rbt{\camera}{\shoe}$. We then crop or pad the image symmetrically so that it is the same size as the original. To implement the shifting augmentation, we shift the image by $i = \frac{\theta}{fov_v} h$  and $j = \frac{\phi}{fov_u} w$ pixels, where $h$ and $w$ are the height and width of $I^g$ in pixels, and $fov_u$ and $fov_v$ are the field of view of the camera in radians in the $u$ and $v$ directions.

    
    
    
    
    
    



When trained on a dataset augmented in this fashion, trajectory prediction becomes significantly more robust, and hook-success increases dramatically across camera poses and shoe poses. To make the case for augmentation more concrete, we report that the network trained without augmentation reliably fails when the shoe position deviates from the center of the image by more than roughly $.1w$ pixels (where $w$ is the width of the image in pixels) or the shoe is closer to or farther away from the camera than the narrow range of $z$ distances present in the training set (roughly 55-65 cm). As evidenced in the experimetns in the following section, \SGNet trained with augmentation performs similarly no matter where the shoe is in the image (as long as it is completely visible) and in a much wider band of $z$ values (roughly 40-100cm in our experiments).
\section{Experiments}\label{sec:experiments}
We perform four experiments in order to motivate our method and characterize its performance on the shoe hooking task. In Experiment 1, we compare the generalization ability of RRT-generated trajectories to that of the learned network when the shoe is varied, demonstrating that as with shoes themselves, there is no {\em one size fits all} shoe-hooking trajectory. In Experiment 2, we characterize on the general level of performance of our learned system when both shoe and shoe pose are varied. In Section \ref{sec:generalizeation-across-pose}, we evaluate the consistency of performance of \SGNet as shoes are moved to different positions in the camera frame, characterizing the effectiveness of our camera-view augmentation strategy in generalizing network performance to shoes not in the center of the frame.  Finally, in Experiment 4, we examine the network's ability to generalize to shoes that are not sitting upright on the table by introducing an artificial roll angle to the shoe's pose.

The experimental setup (Figure~\ref{fig:exp_location}) consists of an RGB-D camera, a table, and a robot equipped with a hook tool. All experiments measure shoe picking performance; a shoe picking attempt is considered successful if the shoe is lifted completely off the table and remains hanging on the hook at the end of the robot motion.

\subsection{Experiment 1: Comparison of Training Trajectories}
\label{sec:training-trajectory-comparison}
%
%
This experiment tests whether qualitatively different hooking motions are actually necessary to hook different shoes.  For the set of four shoes shown in Figure~\ref{fig:rrt_plan} and their four corresponding hooking trajectories generated by the RRT, we test whether each trajectory is able to pick up each shoe.  In each test run, one of the four shoes is placed in a known, fixed location on the table, and one of the four RRT trajectories is executed by the robot at that location to attempt to pick up the shoe.  Each shoe and trajectory combination was run 10 times. For comparison, \SGNet was also run 10 times per shoe with the shoe in the center of the camera frame.

Table~\ref{tab:training-trajectory-comparison} shows the experimental results. As expected, each RRT trajectory succeeds 100\% of the time for the shoe for which it was designed, and is generally unsuccessful at picking other dissimilar shoes, indicating that different trajectories are indeed needed to pick up different shoes.  In contrast, \SGNet succeeds about 90\% of the time on all shoes.
\begin{table}[]
\centering
\begin{tabular}{|l|l|l|l|l|}
\hline
                            & \textbf{Oxford}           & \textbf{Heel}             & \textbf{Pump}             & \textbf{Sneaker} \\ \hline
\textbf{Oxford Trajectory}  & \cellcolor{green}100\%    & 0\%                       & 0\%                       & \cellcolor{green!16}10\%             \\ \hline
\textbf{Heel Trajectory}    & 0\%                       & \cellcolor{green}100\%    & \cellcolor{green!56}80\%  & 0\%              \\ \hline
\textbf{Pump Trajectory}    & 0\%                       & 0\%                       & \cellcolor{green}100\%    & 0\%              \\ \hline
\textbf{Sneaker Trajectory} & \cellcolor{green!53}80\%  & 0\%                       & 0\%                       & \cellcolor{green}100\%            \\ \hline
\textbf{PtPNet Trajectory} & \cellcolor{green!70}90\%  & \cellcolor{green!80}90\%  & \cellcolor{green!80}90\%  & \cellcolor{green!56}80\%             \\ \hline
\end{tabular}
\caption{Experiment 1: Comparison of Training Trajectories. In this table, we show how effective each predefined trajectory and \SGNet are at hooking each type of shoe. In general, the static trajectories only work for the shoe they were generated from, while \SGNet demonstrates good performance on all shoes. that was designed for a given shoe is effective at picking up that shoe, and ineffective at picking up other shoes.}
\label{tab:training-trajectory-comparison}
\end{table}
\subsection{Experiment 2: Shoe Hooking Task Performance}
\label{sec:shoe-picking-task}
%
This experiment tests the overall performance of \SGNet at the shoe picking task.  Once again, each of the four shoes shown in Figure~\ref{fig:rrt_plan} is tested.  The Heel, Pump, and Sneaker were all included in the data used to train \SGNet, while the Oxford shoe was withheld as a test set shoe (\SGNet has never seen this shoe or its corresponding RRT-generated trajectory before).

\begin{figure}[t]
    \centering
    \includegraphics[width=0.4\columnwidth]{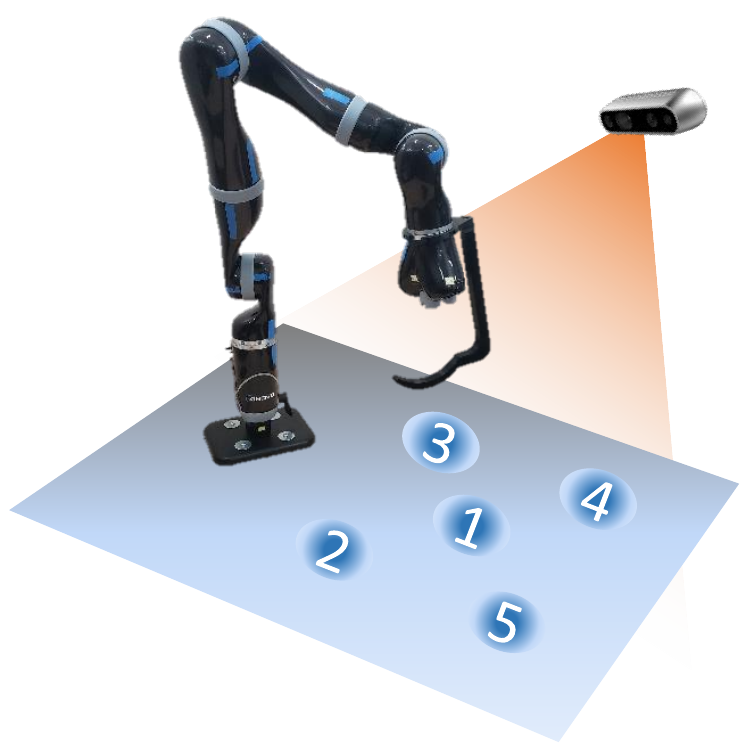}
    \caption{Experimental Setup: Five pre-defined locations on the table. During testing, we placed shoes with a random z-axis rotation at one of these five positions, to maintain consistency across testing runs. \vspace{-0.5cm}}
    \label{fig:exp_location}
\end{figure}

In each trial, the test shoe is placed in a random orientation at one of the five positions on the table shown in Figure~\ref{fig:exp_location}.
A single RGB-D image of the table is captured, segmented, and provided to the network as input. The network outputs a hooking trajectory (sequence of waypoints) in the camera frame $^\camera W_{\hook}$. This trajectory is then transformed to the robot frame via $^\robot T_\camera$, and executed by the robot to attempt to pick up the shoe.  Each shoe is tested in each position about ten times.

\begin{figure}[t]
    \centering
    \includegraphics[width=\columnwidth]{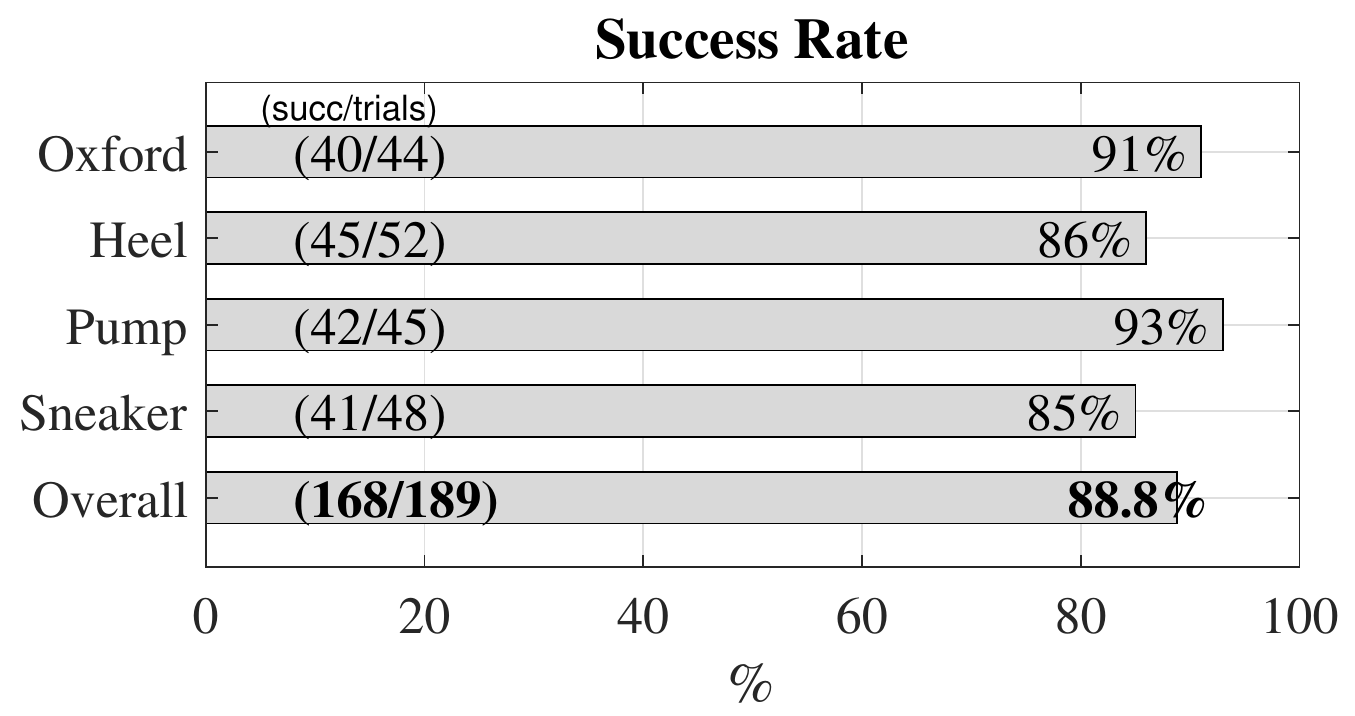}
    \caption{Experiment 2: Overall success rate of \SGNet at the shoe picking task.  The position and orientation of each shoe is varied across the table between trials.  The Oxford shoe was held out of the training set (\SGNet has never seen it before). \SGNet successfully picks up all four of these diverse shoes at least 85\% of the time. \vspace{-0.5cm} }
    \label{fig:shoe-picking-task}
\end{figure}

Figure~\ref{fig:shoe-picking-task} shows the results for each shoe averaged over all trials. The network consistently hooks all shoes at least 85\% of the time, and successfully generalizes to the previously-unseen Oxford shoe, picking it up 91\% of the time.
%
%
\subsection{Experiment 3: Generalization Across Pose}
\label{sec:generalizeation-across-pose}
To test how well the network generalizes to shoes in different positions and orientations within the field of view of the camera, we compare its performance against what we refer to as the `pose estimation system', which explicitly estimates shoe pose and then attempts to pick up the shoe using a human-selected appropriate hooking motion pre-generated by our RRT planner.

To generate picking trajectories, our RRT planner requires a full 3D point cloud of a shoe, as well as a human-specified goal pose for the hook within the shoe.  Consequently, the input to the network (a single-view RGB-D image of a shoe) is not an adequate input to run the RRT planner on live data. Instead, the pose estimation-based system attempts to pick up shoes by estimating the pose of the shoe on the table and then running the (human-selected) pre-computed RRT trajectory for that shoe, taking advantage of the fact that the RRT trajectories are defined with respect to the shoe body frame.  The position of the shoe is estimated by computing the centroid of the (segmented) shoe point cloud in the table frame.  The orientation is estimated by computing the major axis of the shoe point cloud via PCA.

The experimental procedure is the same as the shoe-picking task, except we select a single shoe (the Sneaker), testing only the ability of the pose estimation and learned systems to pick it up when it was placed in each of 5 poses. 

Table~\ref{tab:generalizeation-across-pose} compares the performance of the pose estimation system to the network.  We see that \SGNet slightly outperforms the pose estimation system in most positions, indicating that the {\SGNet}'s  estimation of shoe pose is more accurate than that of the pose estimation system.
\begin{table}[]
\begin{tabular}{|l|l|l|l|l|l|l|}
\hline
\textbf{Position}                                                               & \textbf{1} & \textbf{2} & \textbf{3} & \textbf{4} & \textbf{5} & \textbf{Overall} \\ \hline
\textbf{Network}                                                                & 100\%      & 83\%       & 100\%      & 83\%       & 100\%      & 93.2\%           \\ \hline
\textbf{\begin{tabular}[c]{@{}l@{}}Pose\\ Estimation\end{tabular}} & 83\%       & 100\%      & 100\%      & 50\%       & 67\%       & 80\%             \\ \hline
\end{tabular}
\caption{Experiment 3: Generalization Across Poses. We compare the success rate of two algorithms (\SGNet and a pose estimation-based method) at picking the Sneaker across 5 different starting positions. \SGNet outperforms the pose-estimation method, even though the pose-estimation method automatically uses the correct (human-selected) RRT-generated trajectory to attempt to pick the sneaker.}
\label{tab:generalizeation-across-pose}
\end{table}

%
%
\subsection{Experiment 4: Generalization to Untrained Shoe Orientations}
\label{sec:generalization-across-roll}
%

\SGNet is trained only with images of shoes sitting upright on a flat surface; any variations in shoe roll present in the dataset occur because of the variation in the angle of inclination of the three cameras used. However, our framework is not inherently limited to picking up shoes in these conformations.  In this experiment, we test the ability of the \SGNet to generalize to shoe orientations that are not well represented in the training data.  Specifically, we characterize the sensitivity of performance at the shoe picking task (Sec.~\ref{sec:shoe-picking-task}) when the shoe is resting unevenly on a small block that introduces a rotation about the shoe's heel-toe axis, testing roll angles of $0$, $17$, and $32$ degrees.
Table~\ref{tab:untrained-orientations} shows the results of this experiment. The network successfully picks up shoes with approximately 17 degrees of roll, but fails once the roll angle approaches 30 degrees.
\begin{table}[]
\centering
\begin{tabular}{|l|l|l|l|}
\hline
\textbf{Rotation (Degrees)} & \textbf{0}$^\circ$ & \textbf{17}$^\circ$& \textbf{32}$^\circ$  \\ \hline
\textbf{Success Fraction}   & 41 / 48    & 28 / 36     & 2 / 35      \\ \hline
\textbf{Percentage}         & 85\%       & 78\%        & 6\%        \\ \hline
\end{tabular}
\caption{Experiment 4: Generalization to Untrained Shoe Orientations. The success rate of \SGNet when asked to lift shoes that have been rotated about their principal axis (and thus are no longer perpendicular to the plane). The larger the roll angle, the worse the performance. \vspace{-0.5cm}}
\label{tab:untrained-orientations}
\end{table}
%
%
\section{Discussion}\label{sec:discussion}
The results of our experiments suggest that imitation learning using a kinematic motion planner as a supervision signal is a robust, data-efficient method for single-view estimation of end-effector trajectories for the examined manipulation task, using only 34 human-generated annotations. Further, we posit that the method should generalize to other manipulation tasks in which collision avoidance is important but only partial state observations are available. Our method could also be applied for closed-loop control where the network is applied to subsequent image inputs during the motion execution and used to modulate the initial planned trajectory.


However, the learned system has several clearly defined, repeatable failure modes. For example, we observed that system performance was sensitive to the quality of the segmentation: poor shoe segmentation results often lead the system to failure. There are two primary avenues to addressing this issue, including a) increasing the robustness of \SGNet to bad segmentations and/or b) eliminate the need for masking. Significant progress can be made on both fronts through data augmentation, assuming that the ground truth masks for the dataset image pairs $(I^g,I^d)$ exist. However, segmentations serve as an effective attention mechanism, allowing the network to estimate trajectories for scenes containing multiple objects of interest by selectively masking them. Further, as demonstrated in Experiment 4, the network has only a limited ability to generalize to unseen shoe poses. This is largely due to the range of camera positions used during data collection and could be mitigated by acquiring more complete coverage of the hemisphere of possible camera positions during data collection. In general, when the system fails, it fails because the end-effector collides with the outside of the shoe (a near miss).
%
\section{Conclusion}\label{sec:conclusion}

We presented a self-supervised system which can generate a shoe hooking trajectory directly from a single RGB-D image. Our system is trained using 3D models which are used for generating training instances composed of input images and corresponding hooking trajectories. The hooking trajectories in turn are generated using an RRT planner which takes the 3D model along with goal points as input. The only manual labelling required for our method is the annotation of each of the 34 shoe data bundles (all images and pointclouds from all angles) with a goal point for the calculated RRT trajectories. We also presented a novel augmentation method which was used to generate millions of images over the course of 1000 training epochs from the 7335 training tuples of these 34 data bundles. Hardware experiments demonstrate that the network can successfully hook different types of shoes across a wide range of poses.

Our results suggest several possible paths for future work. A more comprehensive test containing broader manipulation tasks and classes of objects would help better demonstrate the generality of our method. In addition, more experimentation is needed to determine an effective trade-off between neural network architecture size and computational demands; for true real-time trajectory estimation, a more compact architecture backbone is needed. 
\vspace{-0.3cm}


\bibliographystyle{IEEEtran}
\bibliography{references}

\end{document}